\documentclass[letterpaper]{article} 
\usepackage{aaai24}  
\usepackage{times}  
\usepackage{helvet}  
\usepackage{courier}  
\usepackage[hyphens]{url}  
\usepackage{graphicx} 
\usepackage{natbib}  
\usepackage{caption} 
\usepackage{multirow}
\usepackage{amssymb}
\usepackage{pdfpages}
\newcommand{\ie}{\textit{i.e.}}
\newcommand{\eg}{\textit{e.g.}}
\usepackage{algorithm}
\usepackage{color}
\usepackage{booktabs}
\usepackage{pifont}
\usepackage{mathrsfs}

\usepackage{array}
\usepackage{algorithmic}

\usepackage{newfloat}
\usepackage{listings}
\DeclareCaptionStyle{ruled}{labelfont=normalfont,labelsep=colon,strut=off} 
\floatstyle{ruled}
\newfloat{listing}{tb}{lst}{}
\floatname{listing}{Listing}
\title{QAGait: Revisit Gait Recognition from a Quality Perspective}

\author {
    Zengbin Wang\textsuperscript{\rm 1}\thanks{These authors contributed equally to this work.},
    Saihui Hou\textsuperscript{\rm 2,\rm 3}$^{*}$,
    Man Zhang\textsuperscript{\rm 1}\thanks{Corresponding author (zhangman@bupt.edu.cn).},
    Xu Liu\textsuperscript{\rm 3}, \\
    Chunshui Cao\textsuperscript{\rm 3}, 
    Yongzhen Huang\textsuperscript{\rm 2,\rm 3},
    Peipei Li\textsuperscript{\rm 1},
    Shibiao Xu\textsuperscript{\rm 1}
}
\affiliations {
    \textsuperscript{\rm 1}Beijing University of Posts and Telecommunications, 
    \textsuperscript{\rm 2}Beijing Normal University, 
    \textsuperscript{\rm 3}Watrix AI\\
    \{wzb1, zhangman, lipeipei, shibiaoxu\}@bupt.edu.cn, \\ \{housaihui, huangyongzhen\}@bnu.edu.cn, 
    \{xu.liu, chunshui.cao\}@watrix.ai
    \footnotetext[1]{These authors contributed equally to this work.} 
}

\begin{document}

\maketitle

\begin{abstract}
    Gait recognition is a promising biometric method that aims to identify pedestrians from their unique walking patterns. Silhouette modality, renowned for its easy acquisition, simple structure, sparse representation, and convenient modeling, has been widely employed in controlled in-the-lab research. However, as gait recognition rapidly advances from in-the-lab to in-the-wild scenarios, various conditions raise significant challenges for silhouette modality, including 1) unidentifiable low-quality silhouettes (abnormal segmentation, severe occlusion, or even non-human shape), and 2) identifiable but challenging silhouettes (background noise, non-standard posture, slight occlusion). To address these challenges, we revisit gait recognition pipeline and approach gait recognition from a quality perspective, namely QAGait. Specifically, we propose a series of cost-effective quality assessment strategies, including Maxmial Connect Area and Template Match to eliminate background noises and unidentifiable silhouettes, Alignment strategy to handle non-standard postures. We also propose two quality-aware loss functions to integrate silhouette quality into optimization within the embedding space. Extensive experiments demonstrate our QAGait can guarantee both gait reliability and performance enhancement. Furthermore, our quality assessment strategies can seamlessly integrate with existing gait datasets, showcasing our superiority. Code is available at https://github.com/wzb-bupt/QAGait.
\end{abstract}

\section{1\quad Introduction}
Gait recognition has attracted widespread interest in computer vision and biometrics community due to its potential to identify individuals from a distance. 
Unlike other biometric characteristics such as face, fingerprint, and iris, gait is difficult to disguise and can be non-intrusively acquired without any cooperation~\cite{sepas2022deepsurvey}. 
These inherent characteristics contribute to gait recognition a promising solution for wide applications, including surveillance, suspect tracking, motion monitoring, and identification~\cite{wan2018survey, shen2022comprehensivesurvey}.

One notable advantage of gait recognition, as highlighted in previous research~\cite{fan2023opengait, ma2023dynamicAggregate, wang2023dygait}, is its common use of binary silhouette as input. 
This preference stems from its simple structure and sparse representation, which allows researchers to focus on extracting essential motion patterns and action variations, rather than being distracted by background, color, or texture details from RGB images. 
Moreover, this simplified input enables efficient and practical processing. Extensive in-the-lab experiments with shallow models have achieved impressive performance, \ie, 95.9\% on CASIA-B dataset~\cite{hsu2022gaittake} and 92.1\% on OU-MVLP dataset~\cite{dou2023gaitgci}. 

\begin{figure}[t]
    \centering
        \includegraphics[width=0.85\columnwidth]{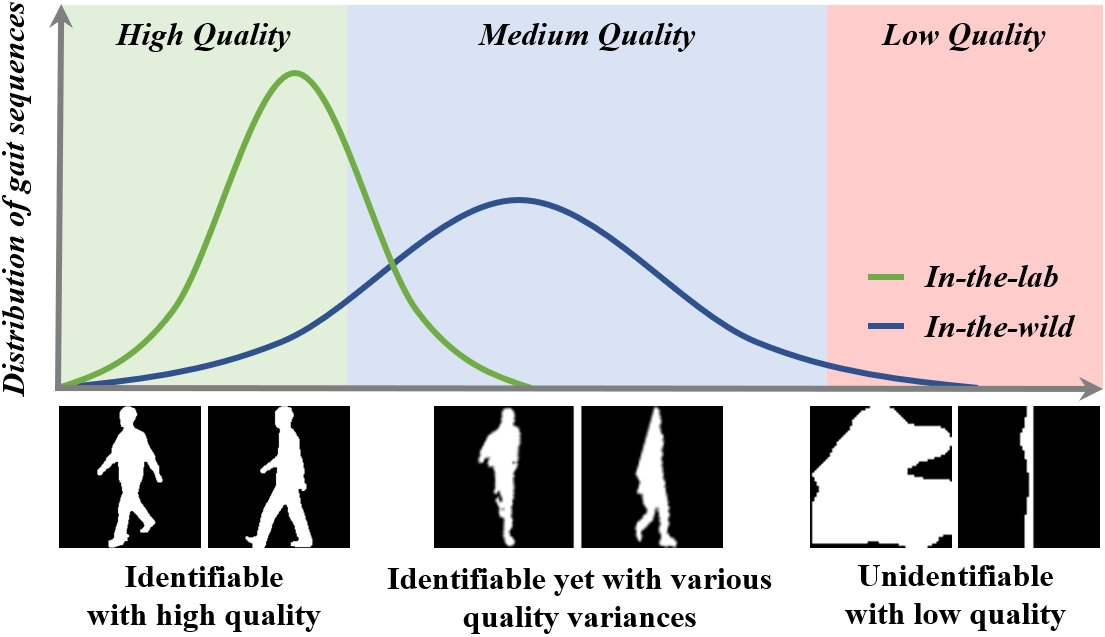}
        \caption{The quality difference when advanced from in-the-lab to in-the-wild gait datasets, including: 1) more extremely low-quality silhouettes that are unidentifiable; and 2) higher quality variances due to various and complex environments.}
    \label{figure01}
\end{figure}

In most cases, the immense success of in-the-lab gait datasets can be attributed to their high-quality segmentation, controlled viewpoint of standard posture, minimal occlusion, \textit{etc}. 
Under these controlled settings, a simple data preprocessing approach~\cite{chao2019gaitset, fan2023opengait} can easily ensure that pedestrians are uniformly processed, thus facilitating subsequent feature extraction and analysis.

However, as in Figure~\ref{figure01}, when gait recognition advances from controlled in-the-lab to in-the-wild scenarios (\eg, Gait3D~\cite{zheng2022gait3d} and GREW~\cite{zhu2021grew}), it encounters various challenges such as complex background, arbitrary camera position, and crowded environment. These uncontrolled conditions lead to two key issues:
\begin{itemize}
    \item \textit{Unidentifiable with low quality}: The traditional data pre-processing approach mentioned above is insufficient to handle the complexities of uncontrolled conditions, potentially reserving a significant number of unexpected challenging silhouettes. Even though some are unidentifiable, all of them are passed to the gait model for feature learning and may potentially affect the model prediction.
    \item \textit{Identifiable yet with various quality variances}: Considerable noise and variability observed across multiple environments increase the uncertainty of gait sequence, resulting in various quality variances. However, current methods usually treat all gait sequences uniformly.
\end{itemize}

These two issues potentially mislead gait model to learn biased gait patterns, thereby resulting in poor performance when compared with in-the-lab settings. 
In this case, a key question arises: Is there a \textbf{\textit{cost-effective quality assessment}} or \textbf{\textit{quality-aware}} approach to mitigate the negative impact of background noises, unidentifiable silhouettes, and identifiable silhouettes yet with various quality variances?

In this paper, we revisit gait recognition from a quality perspective and propose a \textit{unified quality assessment} and \textit{quality-aware} framework, \ie, \textbf{QAGait}, to address the above challenges in a \textit{cost-effective} way. 
Specifically, before feeding into the feature extractor, we propose a series of quality assessment strategies to guarantee silhouette quality, including Maximal Connect Area, Template Match, and Alignment. 
In the feature learning stage, we introduce a gait quality indicator and propose two quality-aware loss functions to adaptively adjust optimization based on inherent gait quality, including Quality Adaptive Margin CE loss (QACE) and Quality Adaptive Margin Triplet loss (QATriplet).

\begin{itemize}
    \item To the best of our knowledge, we make one of the first attempts to explore \textit{quality-oriented} gait recognition, including unified quality assessment strategies and quality-aware feature learning.
    \item Our thoughtful quality assessment strategies provide a unified and cost-effective approach, suitable for both in-the-wild and in-the-lab gait datasets.
    \item Our novel quality-aware feature learning incorporates a reliable gait quality indicator and two quality-aware loss functions, allowing us to dynamically adjust the optimization progress from a quality perspective.
    \item Extensive experimental results on in-the-wild and in-the-lab gait datasets demonstrate the effectiveness of our method. For example, we achieve a remarkable 7.3\% improvement in Rank-1 accuracy on Gait3D dataset. 
\end{itemize}

\section{2\quad Related Work}
\subsection{2.1 Gait Recognition} 
Currently, gait recognition can be primarily categorized into model-based and appearance-based methods.

\textbf{Model-based methods} mostly utilize 2D/3D pose, 3D mesh, and point cloud as input. For example, 
GaitGraph and GaitGraph2~\cite{teepe2021gaitgraph, teepe2022gaitgraph2} utilize human pose estimation to extract 2D poses from RGB images and combine GCN for spatial-temporal modeling. 
SMPLGait~\cite{zheng2022gait3d} employs a 3D SMPL model to learn 3D parameters of body shape, pose, and camera viewpoint as supplementary information. 
LidarGait and LiCamGait~\cite{shen2023lidargait, han2022licamgait} leverage LiDAR to explore point cloud for gait recognition. 
However, all these methods, along with others not explicitly mentioned~\cite{pinyoanuntapong2023gaitmixer, wu2023gaitformer} rely on the prediction from estimation models of pose, mesh, point cloud, and others. 

\textbf{Appearance-based methods} are widely used to learn the inherent spatial and temporal variations in body shape, clothing, and movement dynamics. 
GEINet~\cite{shiraga2016geinet} aggregates each gait sequence to a gait energy image (GEI) for feature learning. 
GaitSet~\cite{chao2019gaitset, chao2021gaitsetPAMI} regards each gait sequence as an unordered set and lots of works~\cite{hou2020GLN, hou2021setresidual, chai2021silhouetteICIP} follow this setting. 
GaitPart~\cite{fan2020gaitpart} proposes to improve part-based feature learning from focal convolution. 
GaitGL~\cite{lin2021gaitgl} develops a 3D global-local feature extractor to ensemble global and local features. 
CSTL~\cite{huang2021cstl} achieves adaptive temporal learning and salient spatial mining. 
MetaGait~\cite{dou2022metagait} proposes to capture omni-scale dependency from spatial/channel/temporal dimensions of gait sequences. 
DyGait~\cite{wang2023dygait} proposes to focus on the extraction of dynamic features. 
Furthermore, other methods such as 3D-Local~\cite{huang20213dlocal}, LagrangeGait~\cite{chai2022lagrange}, GaitMPL~\cite{dou2022gaitmpl}, GaitGCI~\cite{dou2023gaitgci}, and DANet~\cite{ma2023dynamicAggregate} are continuously emerging due to the manifold advantages of using silhouettes, such as easy acquisition, simple structure, sparse representation, and convenient modeling. Meanwhile, as gait recognition moves to real-world scenarios, gait quality becomes crucial, and some researches now require high-quality silhouette~\cite{wang2023landmarkgait}.

\subsection{2.2 Image Quality Assessment}
In computer vision, Image Quality Assessment (IQA) is crucial for evaluating the inherent quality of images. 
Various methods can be broadly categorized into two types: model-based and model-free approaches. 
Model-based methods~\cite{yang2022maniqa, madhusudana2022image, pan2022dacnn} utilize neural networks to automatically learn image quality features. 
Model-free approaches~\cite{al2012comparison, nascimento2023computer, abdelreheem2023zero} usually employ well-designed perceptual metrics such as the Structural Similarity Index (SSIM) and Peak Signal-to-Noise Ratio (PSNR) to quantify image quality. Additionally, various techniques available in libraries like OpenCV can be utilized for image quality enhancement, such as image smoothing, denoising, histogram equalization, \textit{etc}. 

However, quality assessment in gait task is underdeveloped. GQAN~\cite{hou2022gait} attempts to evaluate gait quality using network design at frame and part levels to emphasize interpretability. Differently, our method seeks to mitigate challenges from unidentifiable and variable-quality silhouettes through \textit{quality assessments} and \textit{quality-aware feature learning} without network modification. 

\begin{figure*}[t]
    \centering
        \includegraphics[width=1.4\columnwidth]{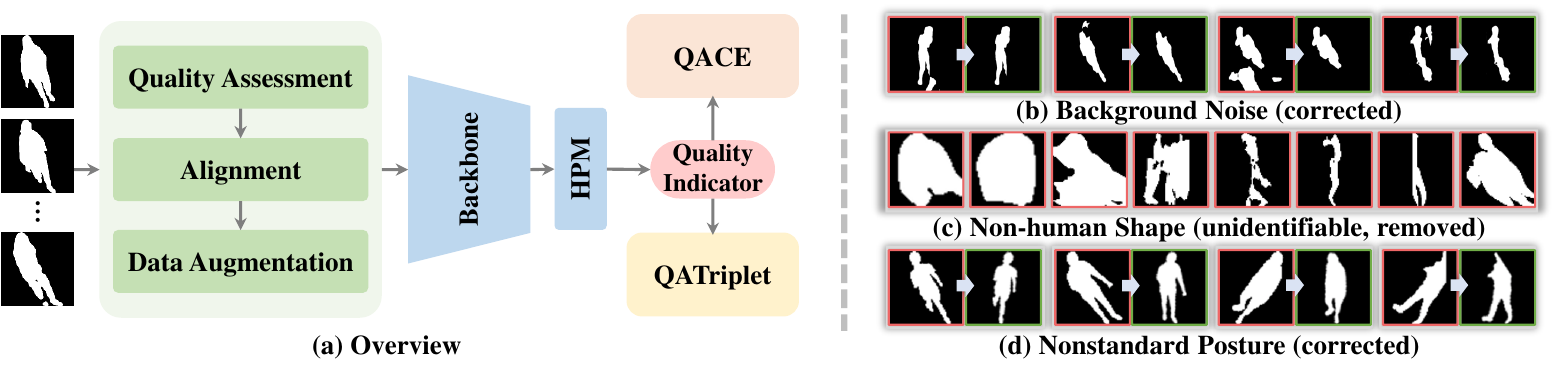}
        \caption{Overview of our QAGait and some examples with quality issues. (a) We design a series of cost-effective quality assessment strategies, alignment, and data augmentation for given gait sequences before feeding them into backbone. During feature learning, we introduce a quality indicator and propose two quality-aware loss functions based on CE loss and Triplet loss (\ie, QACE and QATriplet) to improve model adaptability for various quality variances between gait sequences. (b-d) Some examples with diverse quality issues from in-the-wild gait datasets. HPM: Horizontal Pyramid Mapping~\cite{chao2019gaitset}.}
    \label{figure02:Overview}
\end{figure*}

\section{3\quad Methodology}
In this section, we first revisit the gait recognition pipeline (Section 3.1) and find a series of challenges that are closely associated with silhouette quality. 
%
Then, we present a serious of quality-oriented methods, including Unified Silhouette Quality Assessment (Section 3.3), Alignment (Section 3.4), and Quality-aware Feature Learning (Section 3.5). 

\subsection{3.1 Revisit Gait Recognition Pipeline} \label{Sec:revisit_gait_recognition}

\subsubsection{\textit{1) Early Data Processing.}}
Existing appearance-based gait models often apply a simple data preprocessing approach~\cite{chao2019gaitset, fan2023opengait} to process both in-the-lab and in-the-wild datasets, including:
\begin{itemize}
    \item Crop out black regions at top of head and bottom of feet.
    \item Resize image to the expected height.
    \item Horizontally move human to image center and crop extra side regions to the expected width.
\end{itemize} 

\textbf{[TODO]} However, this process is conducted from early gait recognition and tailored for controlled in-the-lab gait datasets. When transmitting to complex in-the-wild datasets, it is not always suitable for in-the-wild datasets with complex backgrounds, non-human shapes, nonstandard postures, and other unforeseen scenarios.

\subsubsection{\textit{2) Data Augmentation.}} 
Data augmentation in gait recognition is an effective approach to expand the diversity of training dataset, enhance the generalization capability of the model, and prevent overfitting, including: 

\begin{itemize}
    \item Horizontal Flip (HF), Rotation (R), Perspective Transformation (PT), Affine Transformation (AT), Random Erasing (RE).
\end{itemize}

\textbf{[TODO]} However, we note that nonstandard postures are quite common in outdoor datasets~\cite{zheng2022gait3d} due to their unnormalized camera perspectives. This makes using Rotation or Perspective Transformation on an extremely tilted person (\eg, the line connecting the head and feet is nearly parallel to the ground) potentially create even worse patterns that are unidentifiable. 

\subsubsection{\textit{3) Feature Extraction.}}
During the feature extraction stage, the goal is to transform the preprocessed gait sequences into discriminative and compact feature representations. 

Given that this paper does not primarily emphasize the backbone, as a trade-off, we select the latest GaitBase~\cite{fan2023opengait} as our base model for feature representation.

\subsubsection{\textit{4) Loss Function.}} 
In the embedding space, the \textit{softmax cross-entropy loss} and \textit{triplet loss} serve as fundamental loss functions in current gait models~\cite{fan2023opengait}. The former measures the discrepancy between predicted class probabilities and true labels, while the latter pulls the positive pair closer and pushes the negative pair further.

\textbf{[TODO]} However, both of them treat all gait sequences equally and do not account for the significant uncertainty associated with various quality variances in different gait sequences. 
This oversight severely impacts both the convergence speed and its discrimination capability.

\subsection{3.2 Model Overview}
As shown in Figure~\ref{figure02:Overview}, our QAGait presents a comprehensive approach for quality modeling. In quality assessments, we start from Maximal Connect to remove background noise. Only then can we introduce Template Match for accurate shape matching, leaving normal silhouettes for improved Lean-aware Alignment. On top of that, we observe that high quality variance still exists. Thus we further introduce quality-aware losses to mitigate the impact of remaining low-quality silhouettes, which is agnostic to the network and computationally efficient.

\subsection{3.3 Unified Silhouette Quality Assessment}
In this section, as shown in Figure~\ref{figure03:QualityAssessment}(a), we introduce two unified quality assessment strategies, \ie, Maxmial Connect Area and Template Match, aiming to remove the quality issues mentioned above in a \textit{cost-effective} manner. We expect the remaining gait silhouettes can preserve the unique gait patterns, not affected by unrelated factors. 

\subsubsection{\textit{1) Maximal Connect Area.}}
\begin{figure}[t]
    \centering
        \includegraphics[width=1.0\columnwidth]{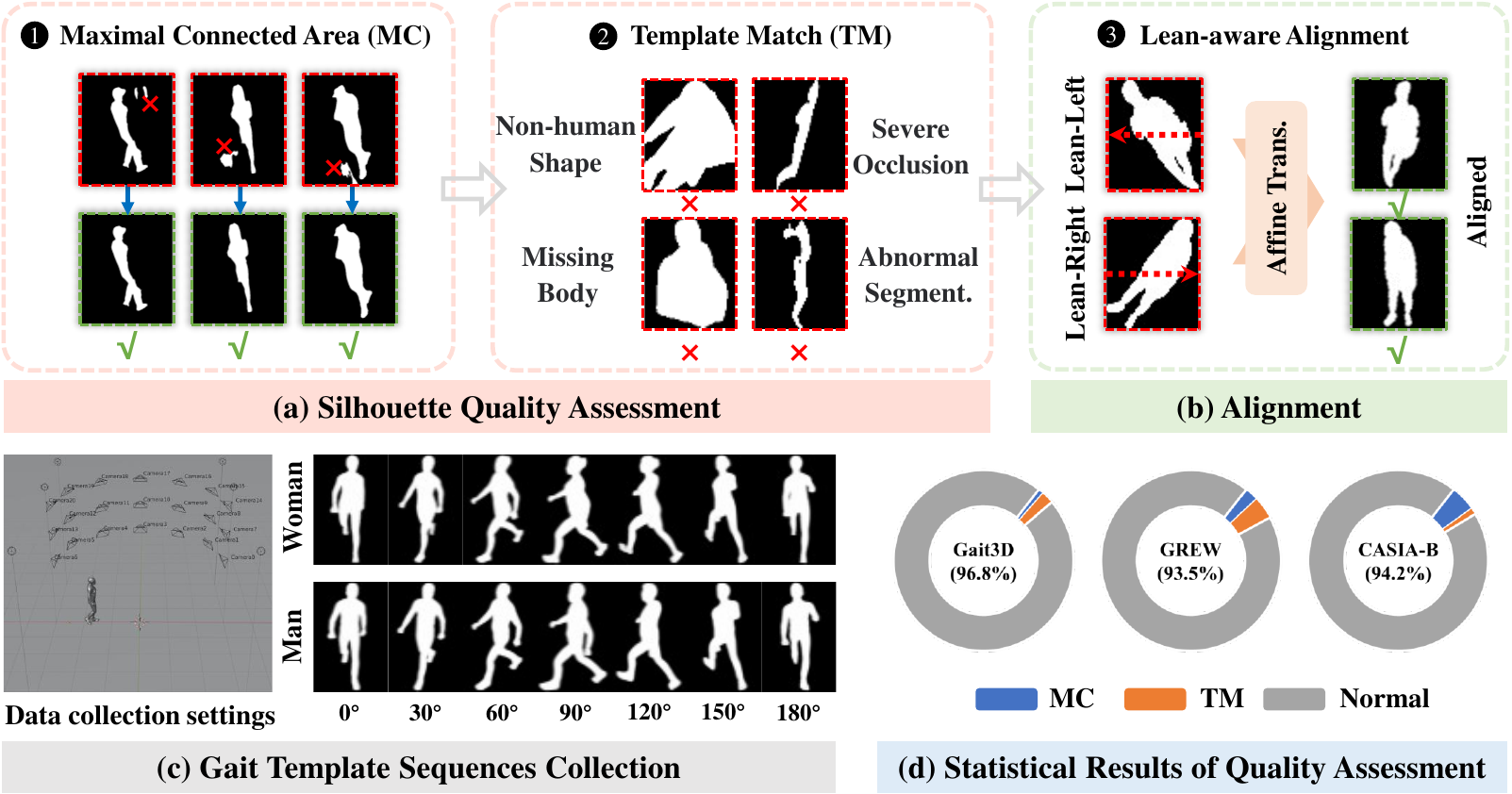}
        \caption{\textbf{Top}: Examples of our silhouette quality assessment and alignment. \textbf{Bottom}: Our gait template sequence collection and the statistical results of quality assessment.}
    \label{figure03:QualityAssessment}
\end{figure}

Since gait recognition is frequently encountered in complex backgrounds and crowded environments, some obvious segmentation errors often exist in outdoor gait datasets. This kind of segmentation error can be largely regarded as background noise and always dissociates from the human body. Thus, as illustrated in Figure~\ref{figure03:QualityAssessment}(a), we develop a Maxmial Connect Area process as our first quality assessment strategy to remove these obvious mistakes.

For details, we calculate the Maximal Connect Area ($area$) by OpenCV (\ie, $cv2.connectedComponents(\cdot)$) for each silhouette, and set a threshold ($\epsilon$) for two cases: 
\begin{itemize}
    \item When $area \geq \epsilon$ and small regions exist, we simply remove these small regions and reserve the main body.
    \item When $area < \epsilon$, the whole silhouette will be removed.
\end{itemize}

\subsubsection{\textit{2) Template Match.}}
As shown in Figure~\ref{figure03:QualityAssessment}(a), only when the first Maximal Connect Area progress removes background noise can we proceed with the second Template Match progress to further remove unidentifiable silhouettes, such as the person with non-human shape, severe occlusion, missing body, or abnormal segmentation.

Building on remarkable processes in 3D SMPL~\cite{loper2023smpl} and improving industry standards in gait recognition~\cite{mirelman2018gait}, leveraging their well-established frameworks makes it possible to create a credible benchmark for standard gait template sequences based on 3D virtual human. 
To this end, as shown in Figure~\ref{figure03:QualityAssessment}(c), we construct an optimal environment and collect a series of standard gait sequences, including various conditions:
\begin{itemize}
    \item Multiple horizontal views: $0^{\circ}$-$180^{\circ}$, with a $30^{\circ}$ interval.
    \item Multiple vertical heights: 1.5 m, 2.5 m, and 3.5 m.
    \item Different genders: woman and man.
\end{itemize}

Even though these standard gait template sequences are available, two challenges still persist for our template match: 
1) Non-normalized camera views in outdoor scenes cause deviations (rotation) from the standard pose; 
and 2) Varying camera distances lead to scale differences. 
Based on these two observations, we employ \textit{Hu Moments}\footnote{Hu moments contain seven distinct invariant moments that exhibit invariance to translation, rotation, and scale changes, providing a compact representation to describe images. This can be efficiently achieved by $cv2.matchShapes(\cdot)$ in OpenCV.}, renowned for their robustness against rotation and scale variations, as our shape-matching algorithm and set a low threshold ($\tau$) to effectively remove non-human shapes. 

\subsection{3.4 Lean-aware Data Augmentation and Alignment}
As mentioned above, we consistently emphasize the nonstandard postures observed in unnormalized camera views. In this case, the commonly used Rotation ($R$) and Perspective Transformation ($PT$) in data augmentation may result in challenging and even unidentifiable samples. Our analysis indicates that current methods lack sensitivity to lean human body. For example:
\begin{itemize}
    \item A person who leans significantly to the left (\eg, $60^{\circ}$ or more) is not suitable to apply leftward rotation. 
\end{itemize}

To address this, we introduce a strict constraint (\ie, lean-aware) in $R$ and $PT$ to prevent excessive deviations. Furthermore, to comprehensively tackle this issue, we propose a lean-aware alignment method aimed at rectifying all silhouettes with nonstandard postures.

\begin{figure*}[t]
    \centering
        \includegraphics[width=1.6\columnwidth]{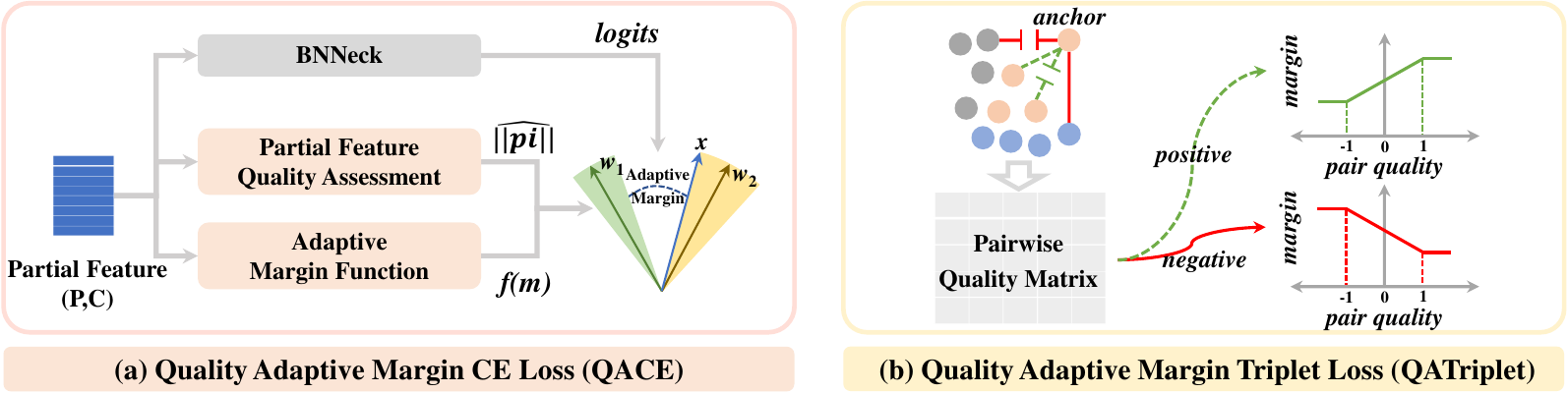}
        \caption{Our quality-aware loss functions facilitate the optimization of trustworthy gait sequences and mitigate the effect of low-quality gait sequences. \textbf{Left}: QACE utilizes Partial Feature Quality as the quality indicator to generate adaptive margin functions. This is then applied to the logits of the model output. \textbf{Right}: QATriplet emphasizes Pairwise Quality as the quality indicator for adaptively adjusting positive and negative pair margins, respectively. BNNeck~\cite{luo2019bag}.}
    \label{figure04:QAloss}
\end{figure*}

\subsubsection{\textit{1) Lean-aware Rotation and Perspective Transformation.}}
To begin with, we uniformly divide the human body into four regions: top-left, top-right, bottom-left, and bottom-right. 
We then compare the calculated area of each region:
\begin{itemize}
    \item If top-left area is larger than top-right by a certain proportion (\eg, 50\%), and bottom-left is smaller than bottom-right by a certain proportion, it is denoted as ``\textit{Lean-Left}'', avoiding leftward rotation. 
    \item The same to ``\textit{Lean-Right}'', without rightward rotation.
\end{itemize}

\subsubsection{\textit{2) Lean-aware Alignment.}}
To fundamentally address the lean body in gait sequences and enable consistent learning, as shown in Figure~\ref{figure03:QualityAssessment}(b), we introduce a lean-aware alignment module, including four steps as follows.
\begin{itemize}
    \item \textbf{Step1}: Find minimal bounding box\footnote{The minimal bounding box is strongly related to the above quality assessment strategies for that background noise and non-human shape will affect the bounding selection and angle calculation. This can be efficiently achieved by $cv2.minAreaRect(\cdot)$.} around human body.
    \item \textbf{Step2}: Calculate rotation angle between the longer side and vertical line.
    \item \textbf{Step3}: Generate affine transformation for correction.
    \item \textbf{Step4}: Add a slight rotation disturbance ($\theta$) to aligned gait sequence for diversity.
\end{itemize}

It is worth noting that a positive or negative angle indicates the lean direction, so we call it Lean-aware Alignment. 
In our implementation, we carefully compare \textit{frame-level} and \textit{sequence-level} alignment and select the latter (\ie, average angles across all silhouettes in a gait sequence.) to ensure consistency within a gait sequence. 

\subsection{3.5 Quality-aware Feature Learning}
In this section, we establish a connection between gait reliability and quality to mitigate various quality variances in remaining gait sequences. Specifically, we introduce a \textit{Quality Indicator} for gait quality assessment and update standard Cross-Entropy loss and Triplet loss toward \textit{Quality Adaptive Margin CE loss (QACE)} and \textit{Quality Adaptive Margin Triplet loss (QATriplet)} for quality-aware feature learning.

\subsubsection{\textit{1) Quality Indicator.}}
Given the pivotal role of partial feature representation in gait recognition (\eg, horizontal strips division to pursue more clues for human body~\cite{chao2019gaitset, fan2020gaitpart}) and the compatibility of feature norm within model-free image quality assessment approaches~\cite{li2018face, schlett2022face, kim2022adaface}, we adopt \textit{Partial Feature Norm} as our quality indicator. 
\begin{equation}
    \fontsize{9pt}{8pt}\selectfont
    \widehat{||p_{i}||} = \left \lfloor \frac{\overline{||p_{i}||}-\mu_{p}}{\sigma_{p}/h} \right \rceil_{-1}^{1}
\end{equation}
where $p_{i}$ indicates partial features extracted from the backbone, $\overline{||p_{i}||}$ indicates average partial features, $\mu_{p}$ and $\sigma_{p}$ denote the batch mean and variance, $\left \lfloor \cdot \right \rceil_{-1}^{1}$ constrains values to facilitate subsequent processing (\ie, $\widehat{||p_{i}||} \in [-1,1]$, from low quality to high quality), $h$ indicates that we follow "$3$-$\sigma$ rule" (also known as the "$68$-$95$-$99.7$ rule") in normal distribution and set $h=0.33$~\cite{lehmann20133}.

\textbf{Discussion.} The quality indicator is a soft manner to integrate quality into loss calculation to mitigate the effect of low-quality frames. 
Its necessity comes from two aspects: 1) Real-world gait data varies in quality, making it hard to adopt a fixed threshold to distinguish high- or low-quality frames easily. Moreover, highly reliable quality metrics are still lacking. 2) To some extent, the low-quality frames (but identifiable) are necessary to ensure the generalization of gait recognition to multiple scenes (\eg, low-resolution).

\subsubsection{\textit{2) Quality Adaptive Margin CE loss (QACE).}}
The softmax cross-entropy loss is widely used for classification and has been advanced to margin-based forms like ArcFace~\cite{deng2019arcface} and CosFace~\cite{wang2018cosface}. The introduction of a margin can compress the learning space to enhance the discriminative capability of the model. Here is the combined form of both as follows.
\begin{equation}
    \fontsize{8pt}{14pt}\selectfont
    L_{1}=-\frac{1}{N}\sum_{i=1}^{N}\log \frac{e^{s(\cos(\theta_{y_{i}} + m_{angle})-m_{add})}}{e^{s(\cos(\theta_{y_{i}} + m_{angle})-m_{add})}+{\textstyle \sum_{j=1,j\neq y_{i}}^{N} e^{s\cos \theta_{y_{i}}}}} 
\end{equation}
where $m_{angle}$ and $m_{add}$ are angular and additive margin for ArcFace and CosFace, $s$ indicates scale transformation, $y_{i}$ is the true labels. The learned embedding features are distributed on a hypersphere with a radius of $s$. 

However, we notice that a fixed margin always treats all features uniformly even when they have varying quality distributions. This will lead to: 
1) Lower-quality gait sequences often cluster near the classification boundary since its mixed and unstable information, and the margin may have a negative effect. 2) Higher-quality gait sequences tend close to the classification center, allowing for a larger margin to compress feature space further.

Therefore, we combine the quality indicator into the margin and dynamically adjust the margin based on the quality of gait sequence. Empirically, we propose to calculate the corresponding $m_{angle}$ and $m_{add}$ as follows.
\begin{equation}
    m_{angle} = -m_{1} \cdot \widehat{||p_{i}||}, \quad m_{add} = m_{1} \cdot \widehat{||p_{i}||}+m_{1}
\end{equation}
where $m_{1}$ is the initial margin and $\widehat{||p_{i}||} \in [-1,1]$ serves as the quality indicator. When $\widehat{||p_{i}||} \rightarrow \{-1\} $, the overall margin is a simple $m_{angle}$ with initial $m_{1}$. When $\widehat{||p_{i}||} \rightarrow \{1\} $, $m_{angle}$ and $m_{add}$ all tend to assign a larger value.

\begin{table*}[t]
    \begin{center}
    \setlength\tabcolsep{11pt}
    \small
    \begin{tabular}{ccccccccc}
    \multicolumn{9}{l}{\textit{\textbf{0. Original Benchmark (Gait3D $|$ GREW)}}} \\
    \toprule [1pt]
    \specialrule{-0.0em}{0.3pt}{0.3pt}
    \multicolumn{1}{c|}{\multirow{2}{*}{Method}} & \multicolumn{1}{c|}{\multirow{2}{*}{Source}} & \multicolumn{3}{c|}{Gait3D $|$ Gait3D$^{\#}$}                                                                                                                                                                                     & \multicolumn{4}{c}{GREW $|$ GREW$^{\#}$}                                                                                                                                                                                                                                  \\
    \multicolumn{1}{c|}{}                        & \multicolumn{1}{c|}{}                        & R-1                                                          & R-5                                                          & \multicolumn{1}{c|}{mAP}                                                          & R-1                                                          & R-5                                                          & R-10                                                         & R-20                                                         \\
    \specialrule{-0.0em}{0.3pt}{0.3pt}
    \midrule
    \specialrule{-0.0em}{0.3pt}{0.3pt}
    \multicolumn{1}{c|}{GaitSet}  & \multicolumn{1}{c|}{AAAI'19}    & 36.7  & 58.3   & \multicolumn{1}{c|}{30.0}  & 46.3  & 63.6  & 70.3  & 76.8  \\
    \specialrule{-0.2em}{0.3pt}{0.3pt}
    \multicolumn{1}{c|}{GaitPart} & \multicolumn{1}{c|}{CVPR'20}    & 28.2  & 47.6   & \multicolumn{1}{c|}{21.6}  & 44.0  & 60.7  & 67.3  & 73.5  \\
    \specialrule{-0.2em}{0.3pt}{0.3pt}
    \multicolumn{1}{c|}{GLN}      & \multicolumn{1}{c|}{ECCV'20}    & 31.4  & 52.9   & \multicolumn{1}{c|}{24.7}  & -     & -     & -     & -     \\
    \specialrule{-0.2em}{0.3pt}{0.3pt}
    \multicolumn{1}{c|}{GaitGL}   & \multicolumn{1}{c|}{ICCV'21}    & 29.7  & 48.5   & \multicolumn{1}{c|}{22.3}  & 47.3  & 63.6  & 69.3  & 74.2  \\
    \specialrule{-0.2em}{0.3pt}{0.3pt}
    \multicolumn{1}{c|}{CSTL}     & \multicolumn{1}{c|}{ICCV'21}    & 11.7  & 19.2   & \multicolumn{1}{c|}{5.6}   & 50.6  & 65.9  & 71.9  & -     \\
    \specialrule{-0.2em}{0.3pt}{0.3pt}
    \multicolumn{1}{c|}{SMPLGait} & \multicolumn{1}{c|}{CVPR'22}    & 46.3  & 64.5   & \multicolumn{1}{c|}{37.2}  & -     & -     & -     & -     \\
    \specialrule{-0.2em}{0.3pt}{0.3pt}
    \multicolumn{1}{c|}{MTSGait}  & \multicolumn{1}{c|}{ACM MM'22}  & 48.7  & 67.1   & \multicolumn{1}{c|}{37.6}  & 55.3  & 71.3  & 76.9  & 81.6  \\
    \specialrule{-0.2em}{0.3pt}{0.3pt}
    \multicolumn{1}{c|}{DAGait}   & \multicolumn{1}{c|}{CVPR'23}    & 48.0  & 69.7   & \multicolumn{1}{c|}{-}     & -     & -     & -     & -     \\
    \specialrule{-0.0em}{0.3pt}{0.3pt}
    \midrule
    \specialrule{-0.0em}{0.3pt}{0.3pt}
    \multicolumn{1}{c|}{GaitBase} & \multicolumn{1}{c|}{CVPR'23}    & 62.4  & 77.7   & \multicolumn{1}{c|}{52.3}  & 58.9  & 73.7  & 79.1  & 83.1  \\
    \specialrule{-0.0em}{0.3pt}{0.3pt}
    \midrule
    \multicolumn{9}{l}{\textit{\textbf{1. After Quality Assessment (Gait3D$^{\#}$ $|$ GREW$^{\#}$)}}} \\
    \midrule
    \specialrule{-0.0em}{0.3pt}{0.3pt}
    \multicolumn{1}{c|}{GaitSet}  & \multicolumn{1}{c|}{AAAI'19}  & 41.7\textcolor{red}{$^{\uparrow5.0}$}  & 61.5\textcolor{red}{$^{\uparrow3.2}$} & \multicolumn{1}{c|}{32.6\textcolor{red}{$^{\uparrow2.6}$}} & 46.5\textcolor{red}{$^{\uparrow0.2}$} & 64.0\textcolor{red}{$^{\uparrow0.4}$}  &70.5\textcolor{red}{$^{\uparrow0.2}$}  &76.9\textcolor{red}{$^{\uparrow0.1}$}   \\
    \specialrule{-0.2em}{0.3pt}{0.3pt}
    \multicolumn{1}{c|}{GaitPart} & \multicolumn{1}{c|}{CVPR'20}  & 31.3\textcolor{red}{$^{\uparrow3.1}$}  & 50.7\textcolor{red}{$^{\uparrow3.1}$} & \multicolumn{1}{c|}{23.9\textcolor{red}{$^{\uparrow2.3}$}} & 46.3\textcolor{red}{$^{\uparrow2.3}$} & 63.2\textcolor{red}{$^{\uparrow2.5}$}  & 69.5\textcolor{red}{$^{\uparrow2.2}$} & 74.4\textcolor{red}{$^{\uparrow0.9}$}  \\
    \specialrule{-0.2em}{0.3pt}{0.3pt}
    \multicolumn{1}{c|}{GLN}      & \multicolumn{1}{c|}{ECCV'20}  & 37.1\textcolor{red}{$^{\uparrow5.7}$}  & 57.4\textcolor{red}{$^{\uparrow4.5}$} & \multicolumn{1}{c|}{29.3\textcolor{red}{$^{\uparrow4.6}$}}  & -                                    & -                                      & -                                     & -                                       \\
    \specialrule{-0.2em}{0.3pt}{0.3pt}
    \multicolumn{1}{c|}{GaitGL}   & \multicolumn{1}{c|}{ICCV'21}  & 34.1\textcolor{red}{$^{\uparrow4.4}$}  & 53.4\textcolor{red}{$^{\uparrow4.9}$} & \multicolumn{1}{c|}{25.1\textcolor{red}{$^{\uparrow2.8}$}}  & 51.4\textcolor{red}{$^{\uparrow4.1}$}&  66.8\textcolor{red}{$^{\uparrow3.2}$}  & 72.4\textcolor{red}{$^{\uparrow3.1}$} &  76.6\textcolor{red}{$^{\uparrow2.4}$}  \\
    \specialrule{-0.0em}{0.3pt}{0.3pt}
    \midrule
    \specialrule{-0.0em}{0.3pt}{0.3pt}
    \multicolumn{1}{c|}{GaitBase} & \multicolumn{1}{c|}{CVPR'23}  & 59.7\textcolor{green}{$^{\downarrow2.7}$} & 77.6\textcolor{green}{$^{\downarrow0.1}$} & \multicolumn{1}{c|}{51.0\textcolor{green}{$^{\downarrow1.3}$}} & 57.2\textcolor{green}{$^{\downarrow1.7}$} & 72.3\textcolor{green}{$^{\downarrow1.4}$} & 77.1\textcolor{green}{$^{\downarrow2.0}$} & 81.3\textcolor{green}{$^{\downarrow1.8}$} \\
    \midrule
    \multicolumn{9}{l}{\textit{\textbf{2. After Quality Assessment \& Quality-aware (Gait3D$^{\#}$ $|$ GREW$^{\#}$)}}}                                           \\
    \specialrule{-0.0em}{0.3pt}{0.3pt}
    \midrule
    \specialrule{-0.0em}{0.3pt}{0.3pt}
    \multicolumn{2}{c|}{QAGait}                        & 67.0\textcolor{red}{$^{\uparrow7.3}$}     & 81.5\textcolor{red}{$^{\uparrow3.9}$}     & \multicolumn{1}{c|}{56.5\textcolor{red}{$^{\uparrow5.5}$}}     & 59.1\textcolor{red}{$^{\uparrow1.9}$}   & 74.0\textcolor{red}{$^{\uparrow1.7}$}     & 79.2\textcolor{red}{$^{\uparrow2.1}$}     & 83.1\textcolor{red}{$^{\uparrow1.8}$}    \\
    \specialrule{-0.0em}{0.3pt}{0.3pt}
    \bottomrule
    \end{tabular}
    \caption{Comparison results on Gait3D and GREW.}
    \vspace{-0.3cm}
    \label{Tab:Gait3D&GREW}
    \end{center}
\end{table*}

\begin{table}[tb]
    \centering
    \setlength\tabcolsep{6pt}
    \small
    \begin{tabular}{ccccc}
    \multicolumn{5}{l}{\textit{\textbf{0. Original Benchmark (CASIA-B)}}}                                                                                                                                                                              \\ 
    \toprule
    \specialrule{-0.0em}{0.3pt}{0.3pt}
    \multicolumn{1}{c|}{Method}       & \multicolumn{1}{c|}{Source}     & NM    & BG    & CL       \\ 
    \specialrule{-0.0em}{0.3pt}{0.3pt}
    \midrule
    \specialrule{-0.0em}{0.3pt}{0.3pt}
    \multicolumn{1}{c|}{GaitSet}      & \multicolumn{1}{c|}{AAAI'19}    & 95.0  & 87.2  & 70.4     \\ 
    \specialrule{-0.2em}{0.3pt}{0.3pt}
    \multicolumn{1}{c|}{GaitPart}     & \multicolumn{1}{c|}{CVPR'20}    & 96.2  & 91.5  & 78.7     \\
    \specialrule{-0.2em}{0.3pt}{0.3pt}
    \multicolumn{1}{c|}{MT3D}         & \multicolumn{1}{c|}{MM'20}    & 96.6  & 92.9  & 82.2     \\
    \specialrule{-0.2em}{0.3pt}{0.3pt}
    \multicolumn{1}{c|}{GaitGL}       & \multicolumn{1}{c|}{ICCV'21}    & 97.4  & 94.5  & 83.6     \\
    \specialrule{-0.2em}{0.3pt}{0.3pt}
    \multicolumn{1}{c|}{3DLocal}      & \multicolumn{1}{c|}{ICCV'21}    & 97.5  & 94.3  & 83.7     \\
    \specialrule{-0.2em}{0.3pt}{0.3pt}
    \multicolumn{1}{c|}{CSTL}         & \multicolumn{1}{c|}{ICCV'21}    & 97.8  & 93.6  & 84.2     \\
    \specialrule{-0.2em}{0.3pt}{0.3pt}
    \multicolumn{1}{c|}{LagrangeGait} & \multicolumn{1}{c|}{CVPR'22}    & 97.5  & 94.6  & 85.1     \\
    \specialrule{-0.2em}{0.3pt}{0.3pt}
    \multicolumn{1}{c|}{MetaGait}     & \multicolumn{1}{c|}{ECCV'22}    & 98.1  & 95.2  & 86.9     \\
    \specialrule{-0.2em}{0.3pt}{0.3pt}
    \multicolumn{1}{c|}{DANet}        & \multicolumn{1}{c|}{CVPR'23}    & 98.0  & 95.9  & 89.9     \\
    \specialrule{-0.0em}{0.3pt}{0.3pt}
    \midrule
    \specialrule{-0.0em}{0.3pt}{0.3pt}
    \multicolumn{1}{c|}{$^{\dag}$GaitBase} & \multicolumn{1}{c|}{CVPR'23} & 96.7  & 93.0  & 75.1    \\
    \specialrule{-0.0em}{0.3pt}{0.3pt}
    \midrule
    \multicolumn{5}{l}{\textit{\textbf{1. After Quality Assessment (CASIA-B$^{\#}$)}}}                    \\
    \midrule
    \specialrule{-0.0em}{0.3pt}{0.3pt}
    \multicolumn{1}{c|}{$^{\dag}$GaitBase}     & \multicolumn{1}{c|}{CVPR'23} & 97.8\textcolor{red}{$^{\uparrow1.1}$} & 94.3\textcolor{red}{$^{\uparrow1.3}$} & 77.1\textcolor{red}{$^{\uparrow2.0}$} \\
    \specialrule{-0.0em}{0.3pt}{0.3pt}
    \midrule
    \multicolumn{5}{l}{\textit{\textbf{2. After Quality Assessment \& Quality-aware (CASIA-B$^{\#}$)}}}                                                                                                                                                      \\ 
    \midrule
    \specialrule{-0.0em}{0.3pt}{0.3pt}
    \multicolumn{2}{c|}{$^{\dag}$QAGait}   & 97.9\textcolor{red}{$^{\uparrow0.1}$} & 94.6\textcolor{red}{$^{\uparrow0.3}$} & 78.2\textcolor{red}{$^{\uparrow1.1}$} \\
    \specialrule{-0.0em}{0.3pt}{0.3pt}
    \bottomrule
    \end{tabular}
    \caption{Comparison results on CASIA-B, ${\dag}$ indicates we reduce the last two layers to suit small-scale CASIA-B.}
    \label{Tab:CASIA-B}
\end{table}

\subsubsection{\textit{3) Quality Adaptive Margin Triplet loss (QATriplet).}}
Triplet loss primarily focuses on pairwise samples, pulling the positive pair closer and pushing the negative pair further apart. The original triplet loss can be simply formulated as:
\begin{equation}
    L_{2} = \max \left \{ d(a,p)-d(a,n)+m_{2}, 0 \right \}
\end{equation}

Similar to QACE, the original triplet loss treats all positive or negative pairs equally. However, we argue that optimizing gait model with lower-quality pairs may not always yield reliable results. For higher-quality pairs, a more effective strategy is to increase the optimization distance for positive pairs while reducing the optimization distance for negative pairs.
\begin{equation}
    L_{3} = \max \left \{ (d(a,p)+m_{ap})-(d(a,n)-m_{an}), 0 \right \}
\end{equation}

We define the pairwise quality indicator ($PQ(s_1,s_2)$) and propose to adaptively adjust the margin ($m_{ap}$, $m_{an}$) based on pairwise quality for an appropriate optimization process.
\begin{equation}
    PQ(s_{1},s_{2}) = \min \left \{ \widehat{||p_{i,s_{1}}||}, \widehat{||p_{i,s_{2}}||}  \right \}
\end{equation}
where $\min$ indicates that if one of gait sequences is unreliable, the overall pairwise quality becomes compromised. 
\begin{equation}
    m_{ap} = m_{an} = 1.5\cdot m_{2} + (0.5\cdot m_{2} \cdot PQ(s_{1},s_{2}))
\end{equation}
where $m_{2}$ is the initial margin, $s_{1}$ and $s_{2}$ are pairwise gait sequences. $m_{ap}$ and $m_{an}$ are personalized margins for positive and negative pairs. The coefficients ($1.5$ and $0.5$) are used to confine the range of $m_{ap}$ or $m_{an}$ within $[m_{2}, 2m_{2}]$. $PQ \in [-1,1]$. As $PQ \rightarrow \{-1\}$, a minimal distance of $m_{2}$ is added, while as $PQ \rightarrow \{1\}$, a maximal distance of $2m_{2}$ is added. These correspond to small and large optimization distances, respectively.

\section{4\quad Experiments}
\subsection{4.1 Experimental Settings}
\subsubsection{Datasets.} 
Our method is primarily evaluated on two in-the-wild gait datasets, Gait3D~\cite{zheng2022gait3d} and GREW~\cite{zhu2021grew}, due to their complex data collection environments and various covariates in outdoor scenes. We also include CASIA-B~\cite{yu2006CASIA-B} since it utilizes outdated background subtraction for segmentation. 
We strictly follow the original training and test settings. 

\subsubsection{Training Details.} 
We adopt the latest GaitBase~\cite{fan2023opengait} as our backbone. 
All data are normalized to $64\times44$. 
We randomly select $P$ identities and corresponding $K$ sequences in a mini-batch (\ie, \{32,4\} for Gait3D / GREW and \{8,16\} for CASIA-B). Each sequence contains 30 randomly sampled silhouettes. 
We use SGD optimizer with weight decay $5e$-$4$. The initial learning rate is $0.1$ and decays $10$ times at each milestone (\ie, \{20K,40K,50K\} for Gait3D / CASIA-B, and \{80K,120K,150K\} for GREW). The total iterations are 60K for Gait3D / CASIA-B and 180K for GREW. 
The overall loss function is: $L=L_{1}+L_{3}$.

\subsubsection{Implementation Details.}
For a fair comparison, we retain at least 15 frames for a sequence when too many frames need to be removed in the quality assessment step. 
For the margin setting, we apply grid search and select the optical ($m_{1}=0.1$, $s=8$) for ArcFace (Table~\ref{Tab:AblationQACE}), and our QACE follows this setting. For QATriplet, we follow the latest gait research and set $m_{2}=0.15$ to achieve the average margin of $m_{ap}$ and $m_{an}$ is around $0.2$.
The threshold\footnote{We invite 20 volunteers, half from our group and half from the college. Each is assigned 30 sequences and asked to select thresholds to distinguish identifiable and unidentifiable frames. Finally, we set the threshold by calculating the \emph{median} of their selections.} in Maximal Connect Area and Template Match is $\epsilon=0.95$ and $\tau=0.001$. 
The random disturbance in Alignment is set as $\theta=5^{\circ}$.

\subsection{4.2 Comparison with State-of-The-Art Methods}
In this section, we evaluate our method on two aspects: 1) original benchmark with our quality assessment (denoted as Gait3D$^{\#}$/GREW$^{\#}$/CASIA-B$^{\#}$), and 2) original benchmark with our whole QAGait (\ie, quality assessment, alignment, data augmentation, and quality-aware feature learning). 

\subsubsection{Evaluation on outdoor datasets (Gait3D \& GREW).}
As shown in Figure~\ref{figure03:QualityAssessment}(d) and Table~\ref{Tab:Gait3D&GREW}, our quality assessment strategies on original Gait3D and GREW can effectively remove unidentifiable silhouettes and reduce noise impact to some extent, \ie, removing 3.2\% / 6.5\% silhouettes for Gait3D / GREW. 
This yields remarkable improvements based on current shallow models. Take GaitPart / GaitGL for example, we achieve 3.1\% / 4.4\% and 2.3\% / 4.1\% Rank-1 gains for Gait3D$^{\#}$ and GREW$^{\#}$ from their original settings. These enhancements align well with our motivation. 

We also adopt the latest GaitBase while it experiences minor drops under default settings (\eg, 62.4\%$\rightarrow$59.7\% in Gait3D$^{\#}$). The probable reason is the mismatch of original data augmentation with the corrected data. The performance can be improved with our data augmentation (59.7\%$\rightarrow$62.2\% in Table~\ref{Tab:AblationAugmentation}). 
Additionally, our cost-effective quality-aware feature learning mitigates various quality issues and achieves remarkable enhancements of 7.3\% and 1.9\% Rank-1 accuracy on Gait3D$^{\#}$ and GREW$^{\#}$ datasets, highlighting the substantial utility of our approach.

\begin{table}[tb]
    \setlength\tabcolsep{6pt}
    \small
    \begin{center}
        \begin{tabular}{l|ccc}
        \toprule
        \specialrule{-0.0em}{0.3pt}{0.3pt}
        \multicolumn{1}{c|}{Conditions} & R-1  & R-5  & mAP  \\
        \specialrule{-0.0em}{0.3pt}{0.3pt}
        \midrule
        \specialrule{-0.0em}{0.3pt}{0.3pt}
        Original Gait3D (w/o DA)        & 55.1 & 72.8 & 46.6 \\
        \specialrule{-0.0em}{0.3pt}{0.3pt}
        \midrule
        \specialrule{-0.0em}{0.3pt}{0.3pt}
        +QA(MaxConnect)                 & 55.8 & 72.9 & 45.7 \\
        \specialrule{-0.2em}{0.3pt}{0.3pt}
        +QA(MaxConnect+TemplateMatch)   & 54.9 & 73.0 & 45.9 \\
        \specialrule{-0.2em}{0.3pt}{0.3pt}
        +QA+DA                          & 62.2 & 78.7 & 51.6 \\
        \specialrule{-0.2em}{0.3pt}{0.3pt}
        +QA+DA+Align                    & 63.5 & 79.8 & 54.2 \\
        \specialrule{-0.2em}{0.3pt}{0.3pt}
        +QA+DA+Align+QACE               & 66.0 & 80.5 & 56.2 \\
        \specialrule{-0.2em}{0.3pt}{0.3pt}
        +QA+DA+Align+QACE+QATriplet     & \textbf{67.0} & \textbf{81.5} & \textbf{56.5} \\
        \specialrule{-0.0em}{0.3pt}{0.3pt}
        \bottomrule
        \end{tabular}
        \caption{Evaluation results on each component in QAGait.}
        \label{Tab:AblationQAGait}
    \end{center}
\end{table}
\begin{table}[tb]
    \centering
    \setlength\tabcolsep{5.5pt}
    \small
    \begin{tabular}{cccccccc}
    \toprule
    \specialrule{-0.0em}{0.3pt}{0.3pt}
    HF        & R               & PT                   & AT   & \multicolumn{1}{c|}{RE}        & R-1           & R-5            & mAP           \\
    \specialrule{-0.0em}{0.3pt}{0.3pt}
    \midrule
    \specialrule{-0.0em}{0.3pt}{0.3pt}
    \ding{51} & \ding{51}       & \ding{51}            &      & \multicolumn{1}{c|}{}          & 59.7          & 77.6           & 51.0          \\
    \specialrule{-0.2em}{0.3pt}{0.3pt}
    \ding{51} & \ding{51}(Lean) & \ding{51}(Lean)      &      & \multicolumn{1}{c|}{}          & 61.1          & 77.2           & 50.8          \\
    \specialrule{-0.2em}{0.3pt}{0.3pt}
    \ding{51} & \ding{51}(Lean) &                      &      & \multicolumn{1}{c|}{\ding{51}} & \textbf{62.2} & \textbf{78.7}  & \textbf{51.6} \\
    \specialrule{-0.0em}{0.3pt}{0.3pt}
    \bottomrule
    \end{tabular}
    \caption{Ablation study on Data Augmentation. ``Lean'' refers to restricting rotation or perspective transformation.}
    \label{Tab:AblationAugmentation}
\end{table}

\begin{table}[tb]
    \centering
    \setlength\tabcolsep{5.0pt}
    \small
    \begin{tabular}{cccccccc}
    \toprule
    \specialrule{-0.0em}{0.3pt}{0.3pt}
    \multicolumn{2}{c}{Frame-level}    & \multicolumn{2}{c}{Sequence-level}  & \multicolumn{1}{c|}{Disturb}   & R-1  & R-5  & mAP  \\
    \specialrule{-0.0em}{0.3pt}{0.3pt}
    \midrule
    \specialrule{-0.0em}{0.3pt}{0.3pt}
    \multicolumn{2}{c}{-}              & \multicolumn{2}{c}{-}               & \multicolumn{1}{c|}{-}         & 62.2 & 78.7 & 51.6 \\
    \specialrule{-0.0em}{0.3pt}{0.3pt}
    \midrule
    \specialrule{-0.0em}{0.3pt}{0.3pt}
    \multicolumn{2}{c}{\ding{51}}      & \multicolumn{2}{c}{}                & \multicolumn{1}{c|}{}          & 63.0 & 78.9 & 53.2 \\
    \specialrule{-0.2em}{0.3pt}{0.3pt}
    \multicolumn{2}{c}{\ding{51}}      & \multicolumn{2}{c}{}                & \multicolumn{1}{c|}{\ding{51}} & 63.0 & 79.1 & 53.7 \\
    \specialrule{-0.2em}{0.3pt}{0.3pt}
    \multicolumn{2}{c}{}               & \multicolumn{2}{c}{\ding{51}}       & \multicolumn{1}{c|}{\ding{51}} & \textbf{63.5} & \textbf{79.8} & \textbf{54.2} \\
    \specialrule{-0.0em}{0.3pt}{0.3pt}
    \bottomrule
    \end{tabular}
    \caption{Ablation study on alignment types and disturbance.}
    \label{Tab:AblationAlignment}
\end{table}

\begin{table}[tb]
    \centering
    \setlength\tabcolsep{5.0pt}
    \small
    \begin{tabular}{ccc|ccc}
    \toprule
    \specialrule{-0.0em}{0.3pt}{0.3pt}
    \multirow{2}{*}{\begin{tabular}[c]{@{}c@{}}Fixed\\ Margin\end{tabular}} & \multirow{2}{*}{\begin{tabular}[c]{@{}c@{}}Quality-aware\\ Margin\end{tabular}} & \multirow{2}{*}{\begin{tabular}[c]{@{}c@{}}Quality\\Calculation \end{tabular}} & \multirow{2}{*}{R-1} & \multirow{2}{*}{R-5} & \multirow{2}{*}{mAP} \\
     &  &  &  &  &  \\ 
     \specialrule{-0.0em}{0.3pt}{0.3pt}
     \midrule
     \specialrule{-0.0em}{0.3pt}{0.3pt}
    - & - & - & 63.5 & 79.8 & 54.2 \\
    \specialrule{-0.0em}{0.3pt}{0.3pt}
    \midrule
    \specialrule{-0.0em}{0.3pt}{0.3pt}
    \ding{51}(0.1) &  &  & 65.0 & 80.2 & 55.6 \\
    \specialrule{-0.2em}{0.3pt}{0.3pt}
     & \ding{51} & Sum-Partial & 65.3 & 80.8 & 55.4 \\
     \specialrule{-0.2em}{0.3pt}{0.3pt}
     & \ding{51} & Min-Partial & 64.0 & \textbf{80.5} & 55.1 \\
     \specialrule{-0.2em}{0.3pt}{0.3pt}
     & \ding{51} & Avg-Partial & \textbf{66.0} & \textbf{80.5} & \textbf{56.2} \\
     \specialrule{-0.0em}{0.3pt}{0.3pt}
     \bottomrule
    \end{tabular}
    \caption{Ablation study on quality-aware margin and quality calculation forms in QACE.}
    \label{Tab:AblationQACE}
\end{table}

\begin{table}[tb]
    \centering
    \setlength\tabcolsep{5.0pt}
    \small
    \begin{tabular}{ccc|ccc}
        \toprule
        \specialrule{-0.0em}{0.3pt}{0.3pt}
        \multirow{2}{*}{\begin{tabular}[c]{@{}c@{}}Fixed\\ Margin\end{tabular}} & \multirow{2}{*}{\begin{tabular}[c]{@{}c@{}}Pair Quality\\ Margin\end{tabular}} & \multirow{2}{*}{\begin{tabular}[c]{@{}c@{}}Pair Quality\\ Calculation\end{tabular}} & \multirow{2}{*}{R-1} & \multirow{2}{*}{R-5} & \multirow{2}{*}{mAP} \\
         &  &  &  &  &  \\
         \specialrule{-0.0em}{0.3pt}{0.3pt}
         \midrule
         \specialrule{-0.0em}{0.3pt}{0.3pt}
        \ding{51}(0.2) &  &  & 66.0 & 80.5 & 56.2 \\
        \specialrule{-0.2em}{0.3pt}{0.3pt}
         & \ding{51} & AvgQualityPair & 63.9 & 80.7 & 54.2 \\
         \specialrule{-0.2em}{0.3pt}{0.3pt}
         & \ding{51} & MinQualityPair & \textbf{67.0} & \textbf{81.5} & \textbf{56.5} \\
         \specialrule{-0.0em}{0.3pt}{0.3pt}
         \bottomrule
    \end{tabular}
    \caption{Ablation study on pairwise quality adaptive margin and pairwise quality calculation forms in QATriplet.}
    \vspace{-0.2cm}
    \label{Tab:AblationQATriplet}
\end{table}

\subsubsection{Evaluation on CASIA-B.}
Since CASIA-B dataset uses outdated background subtraction for segmentation, some segmentation errors may exist. Nevertheless, unlike the Gait3D and GREW datasets, CASIA-B is meticulously collected within a controlled in-the-lab environment, thereby preventing error amplification. Table~\ref{Tab:CASIA-B} shows that our quality assessment yields a modest 2.0\% improvement on CL condition based on GaitBase, and our whole QAGait can further provide an additional 1.1\% improvement.


\subsection{4.3 Ablation Study}

\subsubsection{Effectiveness of different components in QAGait.}
As shown in Table~\ref{Tab:AblationQAGait}, we present our overall approach encompassing quality assessment (QA), alignment, and quality-aware feature learning (QACE \& QATriplet). Our QAGait can progressively improve recognition performance. 

\subsubsection{Impact of strict constraint in data augmentation.}
As shown in Table~\ref{Tab:AblationAugmentation}, lean-aware $R$ and $PT$ can enhance data credibility and lead to a 1.4\% Rank-1 improvement. 
Moreover, our exploration also reveals a new data augmentation combination (\ie, HF, R, and RE) that is better suitable for Gait3D$^{\#}$, with an additional 1.1\% recognition enhancement.

\subsubsection{Impact of align types in alignment.}
We experiment with Frame / Sequence-level Alignment and explore the impact of random disturbance. Table~\ref{Tab:AblationAlignment} shows that sequence-level alignment with a $5^{\circ}$ random disturbance achieves higher performance, indicating that alignment facilitates unified perspective learning and perturbations enhance model diversity.

\subsubsection{Impact of quality adaptive margin in QACE.}
Here we introduce margin into CE loss and develop a quality-aware adaptive margin CE loss (QACE) for evaluation. 
Table~\ref{Tab:AblationQACE} reveals that a fixed margin (\ie, $m=0.1$, similar to ArcFace) enhances Rank-1 accuracy by 1.5\%. 
Moreover, our QACE, with average partial quality, provides an additional performance gain, surpassing the original CE loss by 2.5\%. 

\subsubsection{Impact of pairwise quality calculation in QATriplet.}
Our QATriplet emphasizes pairwise quality of gait sequences and dynamically adjusts optimization distance. Table~\ref{Tab:AblationQATriplet} shows that our QATriplet achieves 1.0\% performance gain with minimizing pairwise quality, highlighting the adaptability to diverse gait sequences.

\subsubsection{Discussion about computation cost.} 
Our approach is computationally efficient largely due to two aspects.
1) The quality assessments are customized for gait based on low-level image processing (OpenCV), making it both affordable and readily available.
2) The quality-aware feature learning is agnostic to the network and non-parametric with only minimal additional computation ($<$1\% training time).
Furthermore, the inference time is unaffected.

\section{5\quad Conclusion}
In conclusion, we present a comprehensive study of gait recognition pipeline and propose a novel quality-oriented gait recognition, namely QAGait. 
Our method involves cost-effective quality assessment strategies and quality-aware feature learning to effectively address challenges arising from unidentifiable silhouettes and identifiable silhouettes yet with various quality variances. 
We expect our method to provide a convenient way to assess gait recognition quality and inspire future research in quality-aware gait recognition.

\section{Acknowledgments}
This work is jointly supported by National Natural Science Foundation of China (62276031, 62276025, 62206022), Beijing Nova Program (Z211100002121106), Shenzhen Technology Plan Program (KQTD20170331093217368), the Fundamental Research Funds for the Central Universities, and Beijing Municipal Science \& Technology Commission (Z231100007423015).

\vspace{.2em}

\bibliography{aaai24}

\newpage
\clearpage

\appendix
\setcounter{table}{0}   
\setcounter{figure}{0}
\renewcommand{\thetable}{A\arabic{table}}
\renewcommand{\thefigure}{A\arabic{figure}}

\subsection{A.1 DataSets}
\subsubsection{Gait3D}
Gait3D~\cite{zheng2022gait3d} is one of the in-the-wild gait datasets. 
It contains 4,000 subjects and 25,309 sequences extracted from 39 cameras in an unconstrained environment. 
These settings bring considerable uncertainty into gait recognition due to factors like non-normalized camera angles, abnormal human postures, random walking sequences, and severe occlusions. 
These challenges encourage a wide exploration of more practical gait recognition.

\subsubsection{GREW}

GREW~\cite{zhu2021grew} is the first in-the-wild gait dataset. 
It consists of 26,000 identities and 128,000 sequences, and all gait sequences are collected from uncontrolled environments. 
The raw videos are captured by 882 cameras in extensive public areas, covering a total of 600 locations. 
Apart from the significant challenges similar to those emphasized in Gait3D, GREW also includes diverse cycling sequences.

\subsubsection{CASIA-B}
CASIA-B~\cite{yu2006CASIA-B} is one of the most popular in-the-lab gait datasets in gait recognition literature. 
It provides 124 identities captured from 11 views (\ie, from $0^{\circ}$ to $180^{\circ}$, with a $18^{\circ}$ interval). 
Each identity contains 10 walking types, including Normal Walking (NM\#01-\#06), Walking with a Bag (BG\#01-\#02), and Walking with a Coat (CL\#01-\#02). 
Since it was collected in 2006 and utilized background subtraction for segmentation, it potentially exhibits various segmentation errors when compared to the latest indoor gait datasets.

\subsection{A.2 Discussion about Template Match Strategy}
Based on our observation of the final results, our Template Match strategy is applicable to various cases in gait task for three reasons. 
\begin{itemize}
    \item \textit{Effective}. Our robust Template Match strategy (Hu Moments) can handle scale variations caused by clothes, and higher-order moments in it are not significantly affected by local variations caused by carrying.
    \item \textit{Sufficient}. The selected templates are largely sufficient to characterize the walking processes in real scenes and those in the existing benchmarks.
    \item \textit{Strict Constraint}. We apply a low threshold ($\tau$) to ensure that we only remove highly non-human and unidentifiable frames in this step.
\end{itemize}

\subsection{A.3 More Visualization of Gait Template Sequence}
\begin{figure}[t]
    \begin{center}
        \includegraphics[width=0.8\columnwidth]{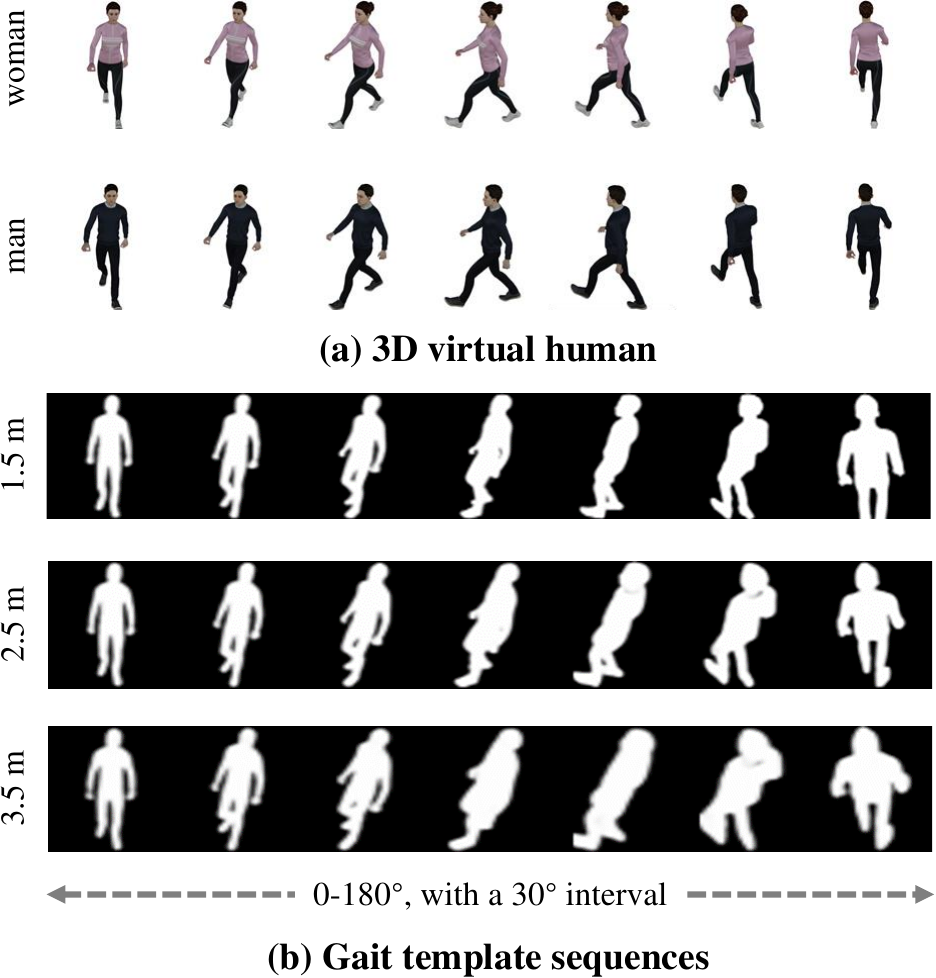}
        \caption{Visualization of our gait template sequences. Each view has 30 frames and here we randomly sample a frame for example. (a) 3D virtual humans, including woman and man. (b) gait templates with different heights and views.}
        \label{supplyfigure01:GaitTemplate}
    \end{center}
\end{figure}

As shown in Figure~\ref{supplyfigure01:GaitTemplate}, we provide more visualization about our proposed gait template sequences. For details, our gait template sequences involve various settings, including:
\begin{itemize}
    \item 7 camera views (0$^{\circ}$-180$^{\circ}$, with a $30^{\circ}$ interval).
    \item 3 heights (1.5 m, 2.5 m, and 3.5 m).
    \item Different genders (woman and man).
\end{itemize}

All gait sequences are conducted by 3D virtual human and each gait template sequence contains 30 frames, with a total of $7\times3\times2\times30=1260$ frames.

\begin{figure*}[t]
    \begin{center}
        \includegraphics[width=1.5\columnwidth]{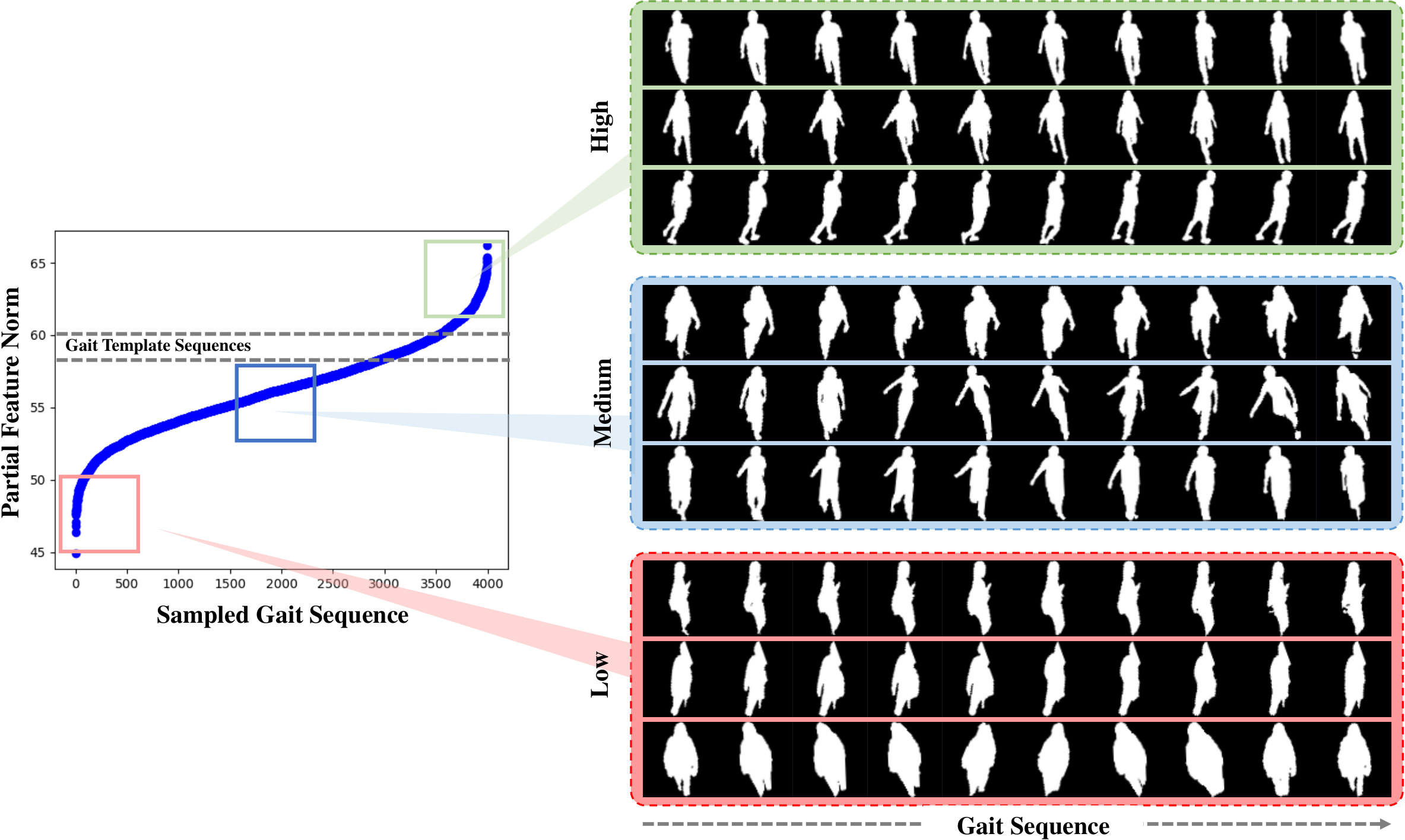}
        \caption{Visualization and analysis of Quality Indicator. We introduce the Partial Feature Norm as the Quality Indicator of gait sequences. \textbf{Left}: An example of quality distribution in outdoor gait dataset. \textbf{Right}: Some examples with low / medium / high Partial Feature Norm, indicating the quality difference of gait sequences.}
        \label{supplyfigure02:PartialFeatureNorm}
    \end{center}
\end{figure*}

\begin{table*}[tbp]
    \setlength\tabcolsep{12pt}
    \small
    \begin{center}
        \begin{tabular}{l|cccc|cccc}
        \toprule
        \specialrule{-0.0em}{0.3pt}{0.3pt}
        \multicolumn{1}{c|}{\multirow{2}{*}{Conditions}}  & \multicolumn{4}{c|}{GREW $|$ GREW$^{\#}$} & \multicolumn{4}{c}{CASIA-B $|$ CASIA-B$^{\#}$} \\
         & R-1  & R-5  & R-10 & R-20  & NM  & BG  & CL & Mean \\
        \specialrule{-0.0em}{0.3pt}{0.3pt}
        \midrule
        \specialrule{-0.0em}{0.3pt}{0.3pt}
        Original DataSet (w/o DA)        & 57.7 & 72.8 & 77.8 & 81.8 & 96.7 & 93.0 & 75.1 & 88.3 \\
        \specialrule{-0.0em}{0.3pt}{0.3pt}
        \midrule
        \specialrule{-0.0em}{0.3pt}{0.3pt}
        +QA(MaxConnect)                 & 55.4 & 70.8 & 76.4 & 80.7 & 98.3 & 94.2 & 77.2 & 89.9 \\
        +QA(MaxConnect+TemplateMatch)   & 55.3 & 70.9 & 76.2 & 80.6 & 97.4 & 93.6 & 76.4 & 89.1 \\
        +QA+DA                          & 57.2 & 72.3 & 77.1 & 81.3 & 97.8 & 94.3 & 77.1 & 89.8 \\
        +QA+DA+QACE               & 58.4 & 73.4 & 78.8 & 82.5 & 97.8 & 94.5 & 77.4 & 89.9 \\
        +QA+DA+QACE+QATriplet     & 59.1 & 74.0 & 79.2 & 83.0 & 97.9 & 94.6 & 78.2 & 90.2 \\
        \specialrule{-0.0em}{0.3pt}{0.3pt}
        \bottomrule
        \end{tabular}
        \caption{Evaluation results of each component in our method. All results are based on GaitBase~\cite{fan2023opengait}. The input image size is $64\times44$ and each gait sequence contains a fixed 30 frames for training. We reduce the last two layers to suit small-scale CASIA-B. $\#$ indicates the updated dataset after our quality assessment strategies. QA: Quality Assessment, DA: Data Augmentation, QACE: Quality Adaptive Margin CE Loss, QATriplet: Quality Adaptive Margin Triplet Loss.}
        \label{SupplyTab:GREWandCASIAB}
    \end{center}
\end{table*}

\subsection{A.4 Visualization and Analysis of Quality Indicator}
In our manuscript, we introduce \textit{Partial Feature Norm} as the \textit{Quality Indicator} of gait sequences to quantitatively represent the various qualities of gait sequences. 
As shown in the right portion of Figure~\ref{supplyfigure02:PartialFeatureNorm}, we present visualizations of gait sequences corresponding to various degrees of \textit{Partial Feature Norm} (\ie, Low, Medium, and High). 
%
\begin{itemize}
    \item Lower values suggest limited gait information in gait sequences, such as standing or occlusion. These cases primarily contain limited shape information rather than diverse motion patterns, which can be considered as low-quality gait sequences.
    \item Medium values maintain the most gait features and motion details, but they can be affected by complex environments, camera angles, or other unseen scenes, resulting in various quality variances.
    \item Higher values are linked to more distinct and comprehensive representations, correlating with higher quality.
\end{itemize}

We also analyze the \textit{quality distribution} of gait sequences within gait datasets. As illustrated in the left portion of Figure~\ref{supplyfigure02:PartialFeatureNorm}, taking Gait3D for example, we randomly sample a gait sequence from each of 4,000 identities, resulting in a total of 4,000 gait sequences. Following this, we compute the values of Partial Feature Norm and sort them in an ascending order. 
By using gait template sequences as the reference, we notice that all gait template sequences have stable and high Partial Feature Norm while the sampled gait sequences in outdoor gait dataset show various quality distributions, indicating a varying degree of quality. 
This observation strongly supports our motivation to develop a quality-aware approach that can effectively account for the diverse quality variances inherent in outdoor gait datasets.

\subsection{A.5 More Results about GREW and CASIA-B}
As shown in Table~\ref{SupplyTab:GREWandCASIAB}, we present the comprehensive experimental results on GREW and CASIA-B. 
Notably, even though our quality assessment strategies sometimes may temporarily decrease model performance, each of them is a strong and cost-effective tool to mitigate the negative effect of unidentifiable silhouettes. 
In this case, the updated gait datasets (\ie, GREW$^{\#}$, CASIA-B$^{\#}$, as well as Gait3D$^{\#}$ in our manuscript) can enhance the credibility and ultimately guide the gait model to learn more trustworthy walking patterns. 
Furthermore, it also can be observed that our proposed module can progressively enhance performance. These results evaluate the effectiveness and rationality of modeling gait recognition from a quality perspective.


\end{document}